\documentclass[lettersize,journal]{IEEEtran}
\usepackage{cite}
\usepackage{amsmath,amssymb,amsfonts}
\usepackage{algorithmic}
\usepackage{graphicx}
\usepackage{textcomp}
\usepackage{xcolor}
\usepackage{multirow,booktabs}
\usepackage{pifont}
\usepackage{hyperref}
\usepackage{soul}
\begin{document}
	\title{Dark Channel-Assisted  Depth-from-Defocus from a Single Image}
	\author{Moushumi~Medhi
	and~Rajiv~Ranjan~Sahay 
	\thanks{M. Medhi is in the Advanced Technology Development Center, Indian Institute of Technology, Kharagpur, India, 721302. e-mail: medhi.moushumi@iitkgp.ac.in}
	\thanks{R. R. Sahay is with department of Electrical Engineering, Indian Institute of Technology, Kharagpur, India, 721302.}}

\maketitle

\begin{abstract}
	We estimate scene depth from a single defocus blurred image using the dark channel as a
	complementary cue, leveraging its ability to capture local statistics and scene structure.  Traditional  depth-from-defocus (DFD)  methods use multiple images with varying apertures or focus. Single-image DFD is underexplored due to its inherent challenges. Few attempts have focused on depth-from-defocus (DFD) from a single defocused image because the problem is underconstrained. Our method uses the relationship between local defocus blur and contrast variations as depth 	cues to improve scene structure estimation. The pipeline is trained end-to-end with adversarial learning. Experiments on real data demonstrate that incorporating  the dark channel prior into  single-image DFD  provides  meaningful depth estimation, validating our approach.	 
\end{abstract}



	\section{Introduction}
\label{sec:intro}
\IEEEPARstart{S}{ingle-image}  depth-from-defocus (DFD) estimates scene depth from a single out-of-focus image.  A single defocused image, captured instantly by a system or robot  without relying on autofocus, can provide fast depth cues.
Blur  from optical limitations can be an advantage, enabling depth extraction where conventional all-in-focus methods  fail.  This paper presents a novel method to estimate depth from a single defocused blurred image captured with a fixed aperture setting.   Existing depth from defocus (DFD) methods  \cite{lin2014extracting,mannan2016discriminative,kumar2018depth,maximov2020focus,lu2021self,song2020multi,si2023fully,wu2024self,fujimura2024deep}  	typically use multiple images captured with varying apertures or focus. These methods exploit the defocus relationship observed among the images with differing focal settings. 	For instance, \cite{lu2021self} jointly trains two networks, DefocusNet and FocusNet, where DefocusNet processes a defocused image to predict depth, which is then used with an input all-in-focus (AIF) image to generate a synthetic focal stack. FocusNet then estimates depth from this focal stack, and its output is combined with the all-in-focus (AIF) image to reconstruct the defocused image. During training, the networks leverage depth and defocus image consistency losses for self-supervision, but at inference, depth estimation can be performed either from a single defocused image or from a focal stack. Unlike these methods, which use video sequences, multiple frames during training or inference, or fuse multiple cues with traditional optimization, we explore a deep learning and dark channel-based
method to address the ill-posed single-image depth-from-defocus (DFD) problem, using only a single
defocused image during training and testing.   This is critical because our approach is designed for scenarios with only a single image, such as monocular systems, making it different and challenging compared to video-based or multi-cue methods.

\begin{figure}[!t]
		\centerline{\includegraphics[height=2.4cm,width=5.5cm]{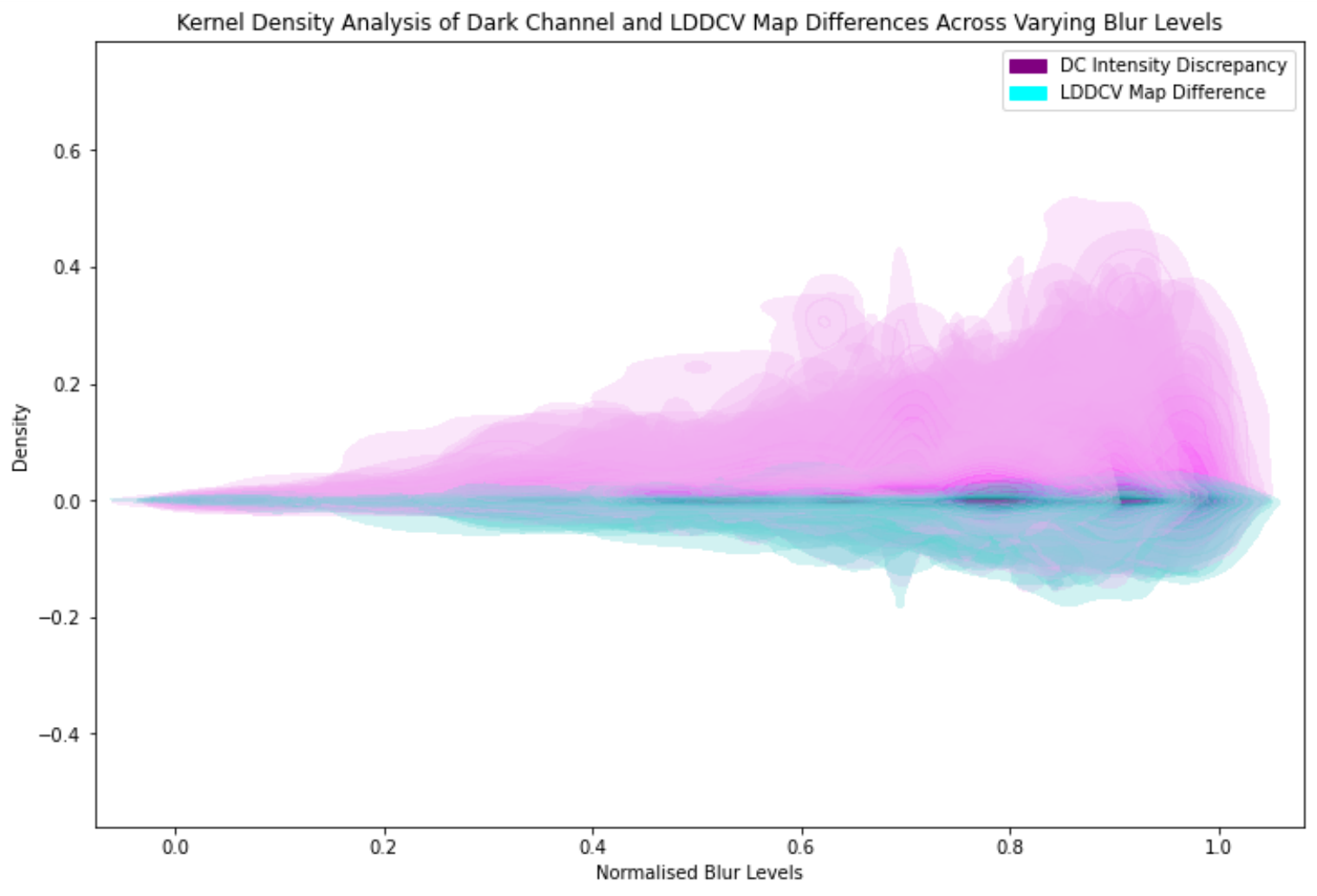}}
		\caption{Kernel Density Estimate Plot of Depth Values in the NYU-Depth V2 Dataset. The KDE plot shows
			the distribution of depth values in the NYU-Depth V2 dataset, generated using a Gaussian kernel with a	bandwidth selected via Silverman's rule.}
		\label{fig:DarkChannelDistriBlur}
	\end{figure}

While multi-image DFD techniques often outperform single-image approaches, single-image DFD remains a significantly more constrained and challenging task. Comparatively, few studies
\cite{anwar2017depth,carvalho2018deep,piche2023lens,wijayasingha2024camera} have addressed depth-fromdefocus (DFD) using a single defocused image, given the problem's difficulty. TThese methods \cite{anwar2017depth,carvalho2018deep,piche2023lens,wijayasingha2024camera} use
end-to-end neural networks to estimate depth maps in a supervised learning setting with ground truth depth
data. To improve depth-from-defocus (DFD) results, \cite{anwar2017depth} also computed blur kernels
for deblurring, while \cite{piche2023lens} derived lens parameters (blur factor and focus disparity) for
defocus blur estimation from the predicted depth map.   \cite{gur2019single} estimates depth from a single AIF image as input and leverages the defocused image solely for supervision during training.  In another depth estimation method from an all-in-focus (AIF) image  \cite{li2021single}, a transmission map, computed from the dark channel, is used as a fourth channel input to a network.   We propose a novel approach to using the dark channel to leverage the relationship between local defocus blur and contrast variations, to deduce the presence and extent of defocus blur, thus providing cues for depth estimation. Dark channel prior (DCP) is commonly used to estimate depth from hazy, foggy, or underwater
images \cite{he2010single,chen2013enhanced,zhou2023underwater}, where DCP is used to compute the scene transmission map, which is a function of depth.  However, dark channel prior (DCP) has also been adapted for space-variant blur analysis for deblurring \cite{yan2017image,pan2017deblurring,cai2020dark} based on dark channel sparsity in deblurred images.
Although defocus blur degradation results from the camera's optics, similar to optical scattering in hazy or foggy conditions, the dark channel plays an analogous role in both types of degraded images.
In defocused blurred images, regions near the focal plane exhibit less blur. The dark channel highlights these regions because of their greater intensity variability. Conversely, the dark channel has reduced intensity variance in significantly blurred areas far from the focal plane and lacks sharp details because of the smoothing effect of blur. We leverage the combined local intensity deviation of the defocused image and its dark channel, namely, 
the Local Defocus and Dark Channel Variation (LDDCV) map, to improve depth-from-defocus (DFD) performance.  The Kernel Density Estimate (KDE) plot for NYU-Depth V2 (NYU-v2) dataset \cite{Silberman2012} in Fig.~\ref{fig:DarkChannelDistriBlur} helps in visualizing how the dark channel intensity discrepancy and the LDDCV map difference change with normalized spatially varying blur level, which is a function of scene depth.
Additionally, we use an adversarial network to supervise our depth-from-defocus (DFD) model, using the defocus blur map as an adversarial signal during training.  Our single-image depth-from-defocus (DFD) approach offers a promising alternative to multiimage or hardware-intensive methods, enabling rapid depth inference from limited data and improving system efficiency. A system could use a fixed-focus, wide-aperture camera (which induces defocus blur) to
passively infer depth from a single shot. This approach reduces system complexity and cost compared to the active depth sensing technique, making it a practical and scalable solution for real-world automation applications. Our empirical results show that applying dark channel prior (DCP) to defocused images yields meaningful depth estimates. 
%
			%
			%
			%
	%
	
	\section{Methodology} 
	\begin{figure*}[!t]
		\centering
		\scalebox{0.9}{
				\begin{tabular}{c }
					
					\includegraphics[height=10cm,width=.9\textwidth]{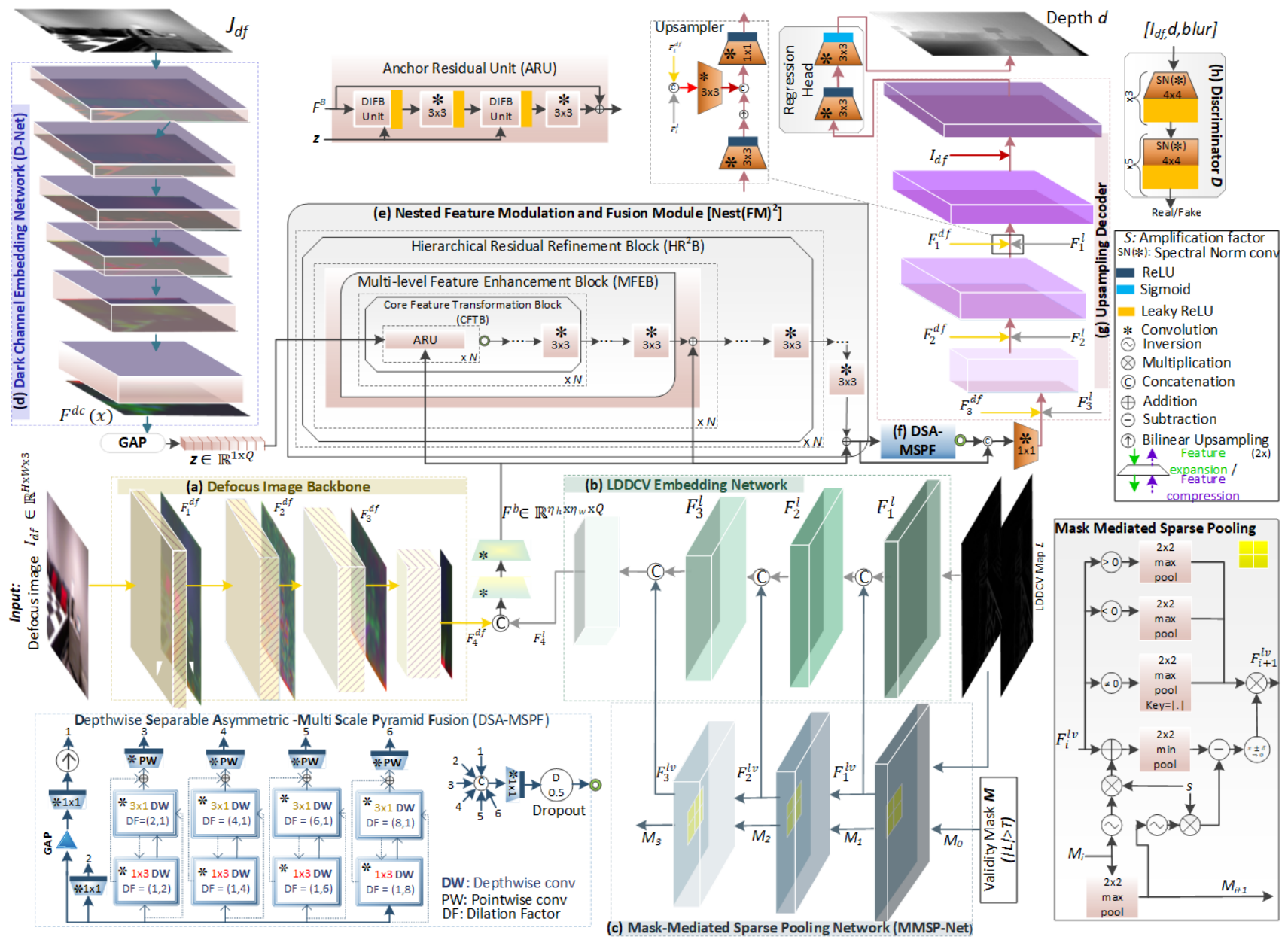}\\
					
			\end{tabular}}
			\caption{Overview of the Dark Channel-Assisted DFD Framework. The diagram illustrates the key modules and workflow of the proposed dark channel-assisted 	DFD framework for image enhancement.}	
			\label{fig:NetArchi1}
		\end{figure*} 
		\begin{figure*}[!t]
			\centering
			\scalebox{0.9}{
					\begin{tabular}{c }
						
						\includegraphics[height=3.5cm,width=.9\textwidth]{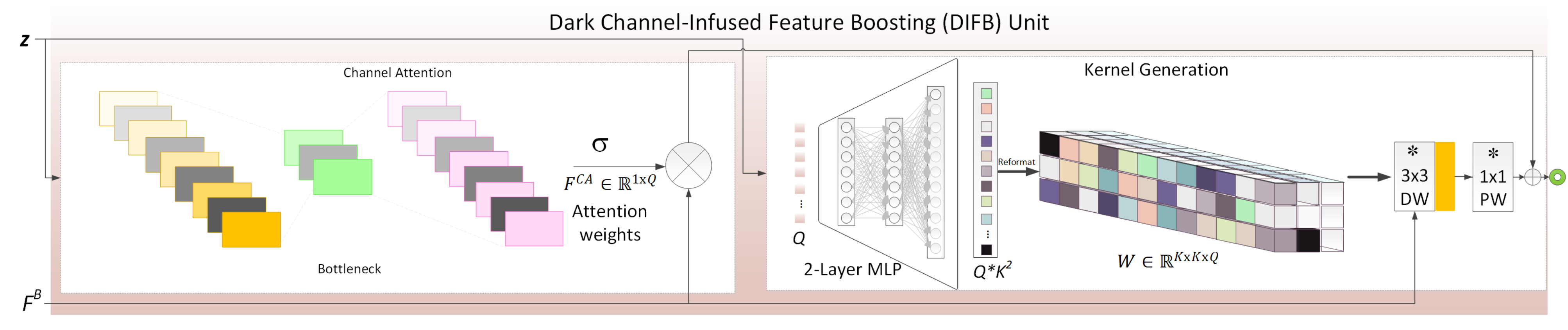}\\
						
				\end{tabular}}
				\caption{Architecture of the Dark Channel-Infused Feature Boosting Unit. The schematic illustrates the
					structure of the DIFB unit, highlighting the integration of dark channel information for feature
					enhancement.}	
				\label{fig:NetArchi2}
			\end{figure*} 
			\subsection{Dark Channel and LDDCV Map as Complementary Cues}
			We compute the darkest scene radiance $ J_{df} $ of the defocused image $ I_{df} $  as the minimum intensity value among the three color channels $ c $ (Red $  r $, Green $ g $, Blue $ b $) in a local window of size $ \Omega (i)\times\Omega (i) $ centered around pixel i of  $ I_{df} $: 
			\begin{equation}\label{eq:darkchanneleqn}
				J_{df}(I_{df})(i)=\min \limits _{p \in \Omega (i)}\left ({\min \limits _{c\in \{R,G,B\}}I_{df}^{c}(p)}\right )
			\end{equation}
			The dark channel emphasizes shadows, edges, and darker structural elements, which can provide
			context for understanding the 3D layout of the scene, such as spatial arrangement and relative distances between objects.  Despite the loss of fine textures and details, the dark channel retains the major scene structure and edges, which correspond to depth transitions. This can improve the clarity of larger structural elements by reducing noise and smoothing out small variations.  We integrated the features extracted from the single defocused image with those of the dark channel to obtain enhanced structural information for the depth estimation model.	 	
			
			TThe LDDCV map is a dual-channel intensity variation map obtained by concatenating the Local
			Defocus Variation (LDV) and the Local Dark Channel Variation (LDCV) maps. They depict the maximum intensity deviation among neighboring pixels within a local region and adequately represent depth-dependent defocus blur.  The LDV and LDCV maps highlight the local variations in $ I_{df} $ and $ J_{df} $, respectively.
			Mathematically,
			\begin{equation}\label{eq:LDDCVeqn}
				\begin{split}
					LDDCV(J, I)(i,j) = \{\max |\, J(i,j) - J(p,q) \,|, \\ \max |\, I(i,j) - I(p,q) \,| \, \Vert \, p = i - 1, i, i + 1, q= j - 1, j, j + 1 \}
				\end{split}
			\end{equation}
  			Defocus blur smooths the image by reducing sharp variations and lowering maximum values within local regions. Because defocus blur homogenizes local regions, areas with high defocus blur show lower local	variations in the local defocus and dark channel variation (LDDCV) map. On the contrary, regions with low-defocus blur show slightly higher LDDCV values. This observation helps determine the presence and extent of defocus blur and provides insights 	to assess the depth of a single out-of-focus image. 
			\begin{figure}[!t]
				\centerline{\includegraphics[height=2.4cm,width=4cm]{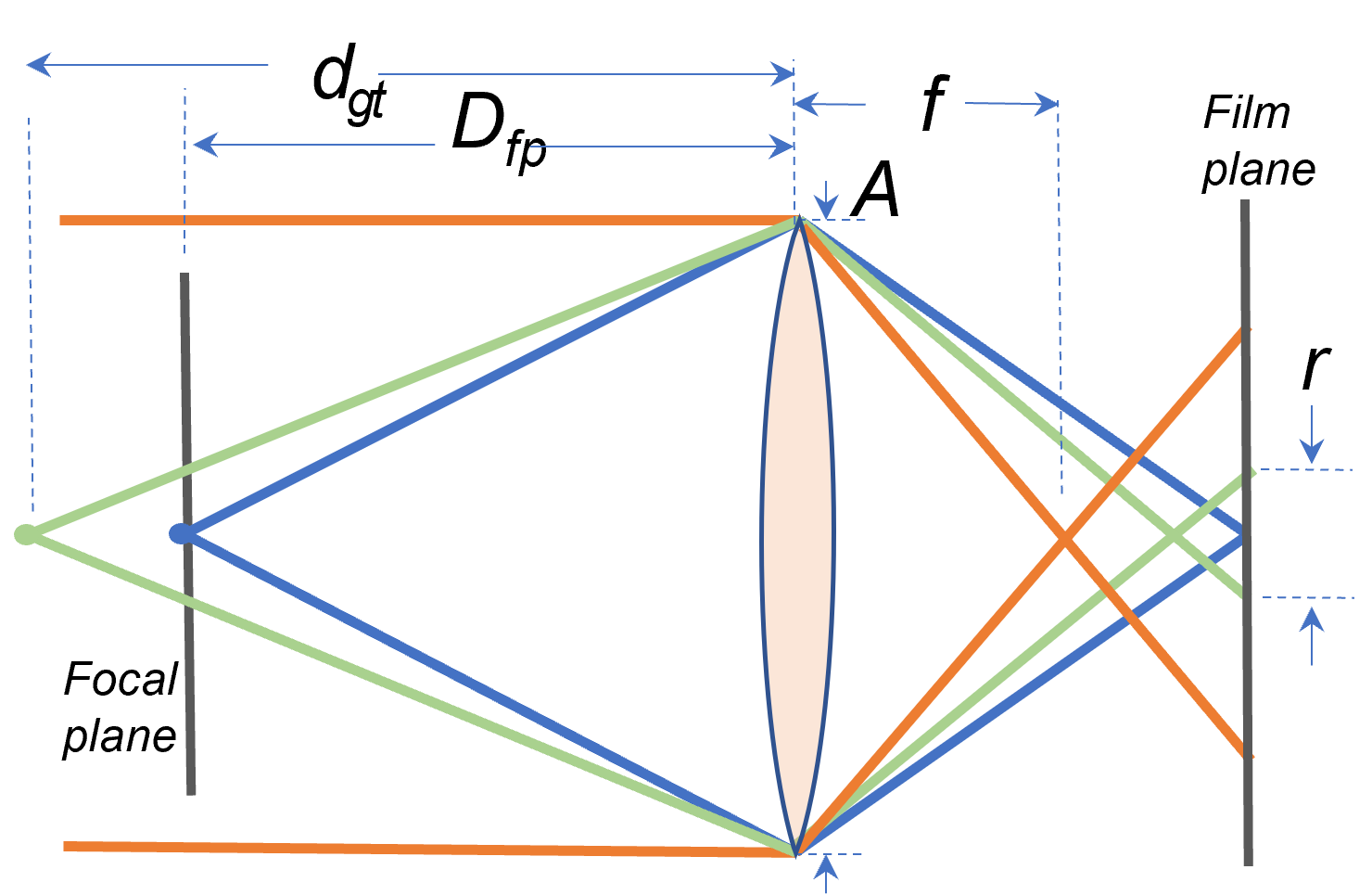}}
				\caption{Thin Lens Model of Defocus Blur. Defocus blur is modeled using
					a thin lens approximation, where the amount of blur is determined by the	distance between the lens and the image plane, as well as the object distance and lens focal length.}
				\label{fig:lens}
			\end{figure}
			
			\subsection{Network Architecture}
			Fig.~\ref{fig:NetArchi1} illustrates our network architecture. 
			For a given defocus image \( I_{df} \in \mathbb{R}^{H \times W \times 3} \), a pretrained ResNeXt101-32x8d-wsl \cite{ranftl2020towards} is employed as the encoder backbone (labeled (a) in Fig.~\ref{fig:NetArchi1}), leveraging the multi-scale features \( F^{df}_{i} \in \mathbb{R}^{H_{i} \times W_{i} \times C_{i}} \) from different encoder layers $i$ ($i=1, 2, 3, 4$).  Here,  $ H_{i} $, $ W_{i} $, and $  C_{i} $ denote the height, width, and channel dimension, respectively. Similarly, multiscale features \( F^{l}_{i} \in \mathbb{R}^{H_{i} \times W_{i} \times C_{i}^{\prime}} \)  are extracted from the LDDCV  embedding network (LDDCV-Net), labeled as (b). Additionally, a parallel  mask-mediated sparse pooling network (MMSP-Net) (labeled (c)) is employed to extract multiscale pooled features \( F^{lv}_{i} \in \mathbb{R}^{H_{i} \times W_{i} \times C_{i}^{\prime\prime}} \)  from the input LDDCV map and its validity mask ($
			1 \text{ if } |\text{LDDCV}| > T
			$,$\text{ where } T = 0.05$ is the threshold), which are then concatenated with \( F^{l}_{i} \).
			The structural information highlighted by the dark channel $J_{df}$ is embedded into a latent
			space (d) by a dark channel embedding network (D-Net), passed through global average pooling (GAP), and flattened to obtain features $z \in \mathbb{R}^{1 \times Q}$.
			The Nested Feature Modulation ($FM$) and Fusion Module ($Nest(FM)^2$), marked as (e), is structured into nested, multi-layered groups to extract  nuanced cues from the embedded dark channel features  $ z $  that modulate the primary features  $ F^{b} \in \mathbb{R}^{\eta_{h_i} \times \eta_{w_j} \times Q} $  in a hierarchical manner. The nested repetition of  ARU, Core Feature Transformation Block (CFTB), Multi-level Feature Enhancement Block (MFEB), and  Hierarchical Residual Refinement ($ HR^2B $) facilitates extensive feature extraction and refinement across multiple levels. The Dark channel-Infused Feature Boosting (DIFB) unit is shown in Fig.~\ref{fig:NetArchi2}. We found that setting N repetitions to 2 achieves a balance between memory efficiency and DFD performance. A subsequent residual module containing a Depthwise Separable Asymmetric-Multiscale Pyramid
			Fusion (DSA-Multiscale Pyramid Fusion (MSPF)) block (marked as (f)) consolidates the learned
			representations by acting as a multi-scale context aggregration prior before passing it to the decoder
			(labeled (g)) for depth, $ d $, reconstruction. 
			We adopted  blueprint separable convolutions (BSConv) \cite{haase2020rethinking} throughout the   depth generator model to reduce our model parameters by approximately \(49\% \).  
			A discriminator $D$ (h) takes the ground truth/estimated depth map ($d_{gt}/d$), ground
			truth/estimated defocus blur map ($r(d_{gt})/r(d)$) (explained in section \ref{sec:dataset}), and the
			defocused image $I_{df}$ as inputs during adversarial training to distinguish between real and generated	data. 
			\subsection{Objective Function}
			The objective function to regress pixel-wise depth values consists of a spatial fidelity loss \( \mathcal{L}_{\text{spafid}} = |d - d_{gt}|_1 \),  a  frequency domain loss \( \mathcal{L}_{\text{freq}} = |\text{Discrete Cosine Transform (DCT)}(d) - \text{Discrete Cosine Transform (DCT)}(d_{gt})|_1 \), and an adversarial loss \( \mathcal{L}_{\text{adv}} = 0.5 \cdot \mathbb{E}_{d \sim p_d} \left[ (D(d, r(d), I_{df}) - 1)^2 \right] \) terms.   The DCT is defined as: \(\text{DCT}(x_i) = x_k = \sum_{i=0}^{L-1} x_i \cos\left[\frac{\pi}{L} \left( i + \frac{1}{2} \right) k \right]\), where $ L $ is the total number of data points in the signal, $ k $ is the index of the DCT coefficients being calculated.   \( \mathbb{E}_{d \sim p_d} \) denotes the expected value over the distribution \( p_d \) of predicted depth map \( d \). The joint loss function is formulated as:
			\begin{equation}
				\mathcal{L}_{\text{total}} = \mathcal{L}_{\text{spafid}} + 0.1\cdot\mathcal{L}_{\text{freq}} + 0.1\cdot\mathcal{L}_{\text{adv}}
				\label{eq:loss1}
			\end{equation}
		\begin{figure*}[!htp]
			\centering
			\scalebox{0.6}{
					\begin{tabular}{c  }
						\includegraphics[height=15cm,width=29cm]{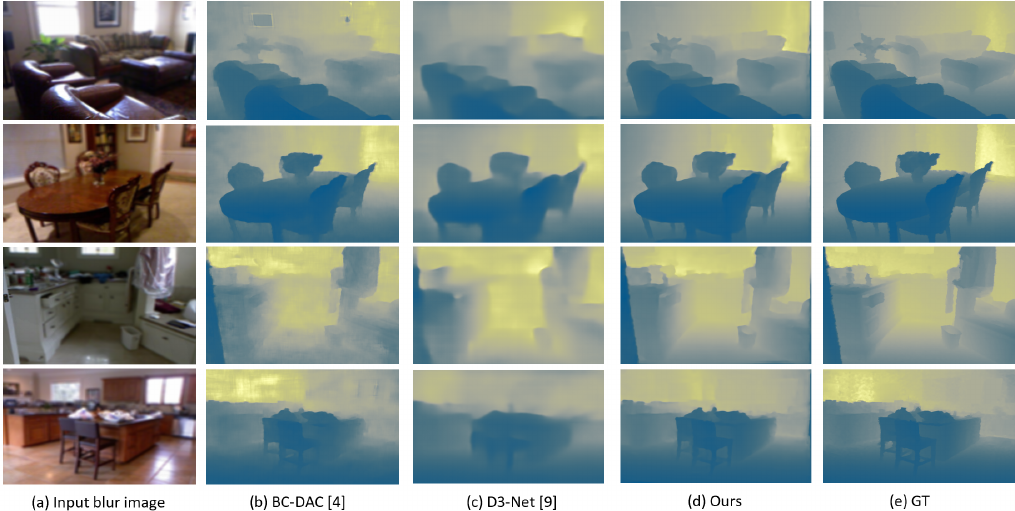}\\
						
				\end{tabular}}
				\caption{Depth Estimation Results on synthesized defocused blur images from the NYU-v2 Dataset. (a) synthesized defocused blur images. (b)-(d) Estimated depth maps. (e) Ground truth depth maps. Images were
				synthesized using data from the NYU-v2 dataset to simulate defocus blur.
				}	
				\label{fig:VISCOMP1}
			\end{figure*}	
			\begin{figure*}[!htp]
				\centering
				\scalebox{0.6}{
						\begin{tabular}{c   }
							
							\includegraphics[height=3.6cm,width=29cm]{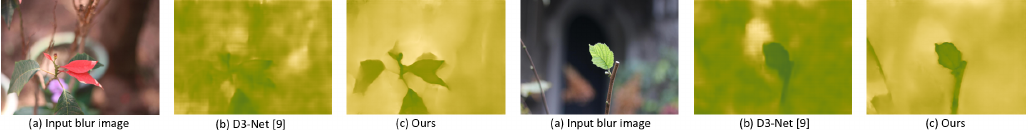}\\
							
					\end{tabular}}
					\caption{Depth estimation on high-resolution real defocused blur images from the EBD Dataset without
						fine-tuning. (a) Depth estimation results for two high-resolution real defocused blur images from the EBD
						dataset, obtained without fine-tuning the model.}	
					\label{fig:VISCOMP2_EBD_DATASET}
				\end{figure*}	
			\section{Experiments and Results}
			
			\subsection{Dataset} \label{sec:dataset}
			\noindent \textbf{NYU-Depth V2 (NYU-v2) dataset \cite{Silberman2012}: } The NYU-v2 dataset
			\cite{Silberman2012} comprises 1,449 pairs of spatially matched Red Green Blue (RGB) and depth images	acquired using a Microsoft Kinect. Following prior work  \cite{carvalho2018deep,song2020multi}, we use the standard train/test		eigen split consisting of 795/654 images.
			To generate optically realistic depth-dependent defocus effects in the all-in-focus (AIF) NYUv2 RGB image $I$, we select parameters corresponding to a synthetic camera with a focal length ($f$) of 9 mm, an in-focus plane ($D_{fp}$) at 0.7 m, an F-number ($F_n$) of 2 to achieve a shallow depth of field (DoF), a sensor size $p_x$ of 7.5 µm, and an aperture $A = f/F_n$.  We generate the defocus-blurred image $I_{df}$ by convolving the all-in-focus (AIF) image $I$ with a point spread function (PSF) $G(x, y, r)$ with kernel radius $r$ and location indices $x$ and $y$:
			\begin{equation}
				I_{df}(x,y) = G(x,y) * I(x,y) \text{, } G(x, y) = \frac{1}{2\pi r^2} e^{-\frac{1}{2} \frac{x^2 + y^2}{r^2}}
				\label{eq:data1}
			\end{equation}
			Following the thin-lens model in Fig.\ref{fig:lens}, $r$ is calculated as a function of the	scene distance, $d_{gt}$, from the camera:
			\begin{equation}
				r(d_{gt}) = \frac{1}{\sqrt{2}\cdot p_x}\frac{Af}{(D_{fp} - f)}\frac{|d_{gt} - D_{fp}|}{d_{gt}}
				\label{eq:data2}
			\end{equation}
	
				\noindent \textbf{Enhanced Blur Dataset (EBD) dataset \cite{jin2023depth}: } The Enhanced Blur Dataset
				(EBD) dataset \cite{jin2023depth} contains 1,305 high-resolution (1600$\times$1024) real defocused images	without ground-truth depth map annotations.  These images feature a shallow depth of field (DoF) with an $F_n$ of 1.8. Note that we have used the EBD dataset \cite{jin2023depth} solely for testing.
				
			\begin{table}[!t]
				\caption{Quantitative depth estimation results on the NYU-v2 dataset. This table shows quantitative
					results of depth estimation on the NYU-v2 dataset, with $S_{blur}$ indicating training supervision from a
					defocus blur map. Bold entries indicate the best performance.
				}	
				\begin{center}
					\scalebox{.6} 
					{
						\begin{tabular}{c|c|c  c c c c c c }
							\toprule \hline
							Methods& $S_{blur}$ &	Abs Rel $\downarrow$ &Sq Rel $\downarrow$ & RMSE [m] $\downarrow$  & $log_{RMSE}$ $\downarrow$  &$\delta_{1}$$ \uparrow $ & $\delta_{2}$$ \uparrow $ & $\delta_{3}$$ \uparrow $\\ 
							\hline	\bottomrule 
							\multicolumn{9}{l}{\textit{All-In-Focus (AIF) image(s)}} \\
							\bottomrule \hline 	
							P3Depth\cite{patil2022p3depth}& No &	0.104   &- & 0.356  &- & 0.898& 0.981& 0.996\\ 
							Marigold\cite{ke2024repurposing}& No &	\textbf{0.055} &- & \textbf{0.224} & -  &\textbf{0.964}&	\textbf{0.991}&\textbf{	0.998}\\ 
							
							\hline	\bottomrule 
							\multicolumn{9}{l}{\textit{Focal stack}} \\
							\bottomrule \hline 	
							
							SSDC\cite{si2023fully}& Yes &	0.170  &  - & 0.325  &   - & 0.950 &0.979& 0.987  \\

							\hline	\bottomrule 
							
							\multicolumn{9}{l}{\textit{Dual defocused images}} \\
							\bottomrule \hline 	
							
							BC-DAC\cite{song2020multi} & No &	0.026 &0.007 & 0.140  & 0.018  & 0.995 & 0.998  & 0.999\\

							\hline	\bottomrule 
							\multicolumn{9}{l}{\textit{Single defocused image}} \\
							\bottomrule \hline 	
							
							D3Net\cite{carvalho2018deep} & No &	0.104 &0.056 & 0.384 & 0.057  & 0.923 &  0.987  & 0.996\\
							Camind\cite{wijayasingha2024camera} & Yes &	0.242 & 0.248  & 0.798 & 0.253  & 0.601 & 0.917 & 0.990\\
							
							Ours & Yes &\textbf{	0.042} &\textbf{ 0.019} & \textbf{0.240} & \textbf{0.032}  & \textbf{0.975} &  \textbf{0.995} &  \textbf{0.999}\\

					\end{tabular}}
					\label{tab:nyusyn}
				\end{center}
			\end{table}

					\subsection{Quantitative and Qualitative Results}
					To quantitatively evaluate the depth
					estimation results, we show comparison in Table~\ref{tab:nyusyn} for 4 categories of input methods in terms of  7 metrics  that are widely used for depth estimation: Absolute Relative Error ($ Abs Rel $), Square Relative Error ($ Sq Rel $), Root Mean Squared Error ($ RMSE $), logarithmic Root Mean Squared Error ($ log_{RMSE} $), and thresholded accuracies ($\delta_{1}<1.25$, $\delta_{2}<1.25^2$, $\delta_{3}<1.25^3$). For fair comparison, we trained and tested D3Net\footnote{\url{https://github.com/marcelampc/d3net\_depth\_estimation}} \cite{carvalho2018deep} and Camind\footnote{\url{https://github.com/sleekEagle/defocus_camind.git}} \cite{wijayasingha2024camera} using our dataset. We would like to mention here that in contrast to the original work in \cite{wijayasingha2024camera}, which reported test results on NYU-v2 data within a distance range of 2 m, our evaluation included the full range (10 m) of the data. This could account for the performance discrepancy, possibly explaining the lower error rates the authors reported in their paper. The results reported for BC-DAC \cite{song2020multi} in Table~\ref{tab:nyusyn} and Fig.~\ref{fig:VISCOMP1} were provided by the authors. Our method outperforms existing methods that use single defocus input \cite{carvalho2018deep,wijayasingha2024camera}, focal stack input \cite{si2023fully}, and all-in-focus (AIF) input \cite{patil2022p3depth,ke2024repurposing} in most evaluation metrics.  Fig.~\ref{fig:VISCOMP1} shows the  qualitative results. While the outputs of BC-DAC \cite{song2020multi} exhibit noticeable artifacts, the method occasionally produces more accurate depth values, particularly at greater distances (fourth row in Fig.~\ref{fig:VISCOMP1}). This may explain the superior quantitative results shown in Table~\ref{tab:nyusyn}. Overall, our method generates visually meaningful results.
					
					We also evaluate the zero-shot generalization capability of our method on real defocused EBD  data \cite{jin2023depth} with entirely different blur magnitudes and extents that were not   encountered during training, as shown in Fig.~\ref{fig:VISCOMP2_EBD_DATASET}. Unlike D3Net \cite{carvalho2018deep}, the trained model were not fine-tuned on the new dataset. We assume that fine-tuning our model would naturally improve the results on the real dataset. Figs.~\ref{fig:VISCOMP2_EBD_DATASET} (b), (c), (e), and (f) show that our trained model	produces reasonably accurate and more generalizable zero-shot results than D3Net.

					\subsection{Ablation Studies}
					
					\begin{table}[!t]
						\caption{ Ablation study results. This table presents the results of an ablation study, evaluating the
							impact of different components. DCC (dark channel as a complementary cue) and ADV (adversarial learning)
							represent the specific components ablated. Best performing results are indicated in bold. }
						\begin{center}
							\scalebox{0.7}{
								
								\begin{tabular}{c|c c|c c c c c}
									\toprule
									&DCC  & ADV & Abs Rel $\downarrow$ & RMSE$\downarrow$  &  $\delta_{1}$$ \uparrow $ & $\delta_{2}$$ \uparrow $ & $\delta_{3}$$ \uparrow $\\
									\midrule
									
									\ding{172} &            &     \checkmark    & 0.118 & 0.421  & 0.825&	0.980&	0.995  \\
									\ding{173} & \checkmark &     \checkmark    & 0.077 & 	0.362 &  	0.937&	0.984&	0.995 \\
									
									\midrule
									\ding{174} & \checkmark &          & 0.066& 0.287 &  	0.966&	0.992&	0.998    \\
									
									\midrule
									\ding{72} & \multicolumn{2}{c|}{Ours (full)}  & \textbf{0.042} &\textbf{ 0.240}  &\textbf{0.975} &  \textbf{0.995} &  \textbf{0.999}  \\
									\bottomrule
									
								\end{tabular} 
							}
							\label{tab:abl}
						\end{center}
					\end{table}
					We report the  ablation results on NYU-v2 test data in  Table \ref{tab:abl}. The model without the use of dark channel as a complementary cue (DDC)  yields the least impressive results (\ding{172}). Introducing DDC (\ding{173}) into the model by propagating the concatenated dark channel and defocus RGB image as a 4-channel input through the image encoder, while retaining the LDDCV-Net and MMSP-Net, marks an uptick in performance. In this configuration, D-Net and $Nest(FM)^2$  are excluded. Training without adversarial supervision ($\\ding{174}$), i.e., without the discriminator,	slightly degrades performance compared to our full model ($\\ding{72}$).
					\section{Conclusion}
				We presented a novel method to infer depth from a single space-variant defocused image.  We have investigated the
				influence of dark channel and its local intensity variation as
				guidance based on their blur representational features for depth estimation. We introduced the Local Defocus and Dark Channel Variation (LDDCV) map as complementary cues to capture spatial blur cues and local intensity deviations, enabling more accurate depth inference. Additionally, we incorporated adversarial training with defocus blur maps as supervisory signals to improve the quality and realism of the predicted depth maps. Experiments on a realistic synthetic dataset and real defocused data show the potential of our	method. Our findings suggest that the dark channel, traditionally used in dehazing and deblurring tasks, can serve as a meaningful and reliable indicator of local scene structure in defocused images. While promising, our approach has certain limitations. In particular, reliance on synthetic training data may hinder generalization in highly dynamic or cluttered real-world environments. To mitigate this, exposing the model to a broad range of variations across both synthetic and real settings can enhance its robustness and adaptability.

\bibliographystyle{IEEEtran}
\bibliography{DFD_BIB_abbrev}

\begin{thebibliography}{10}
\providecommand{\url}[1]{#1}
\csname url@samestyle\endcsname
\providecommand{\newblock}{\relax}
\providecommand{\bibinfo}[2]{#2}
\providecommand{\BIBentrySTDinterwordspacing}{\spaceskip=0pt\relax}
\providecommand{\BIBentryALTinterwordstretchfactor}{4}
\providecommand{\BIBentryALTinterwordspacing}{\spaceskip=\fontdimen2\font plus
\BIBentryALTinterwordstretchfactor\fontdimen3\font minus
  \fontdimen4\font\relax}
\providecommand{\BIBforeignlanguage}[2]{{%
\expandafter\ifx\csname l@#1\endcsname\relax
\typeout{** WARNING: IEEEtran.bst: No hyphenation pattern has been}%
\typeout{** loaded for the language `#1'. Using the pattern for}%
\typeout{** the default language instead.}%
\else
\language=\csname l@#1\endcsname
\fi
#2}}
\providecommand{\BIBdecl}{\relax}
\BIBdecl

\bibitem{lin2014extracting}
X.~Lin, J.~Suo, and Q.~Dai, ``Extracting depth and radiance from a defocused
  video pair,'' \emph{IEEE Trans. Circuits Syst. Video Technol.}, vol.~25,
  no.~4, pp. 557--569, 2014.

\bibitem{mannan2016discriminative}
F.~Mannan and M.~S. Langer, ``Discriminative filters for depth from defocus,''
  in \emph{Int. Conf. 3D Vis. (3DV)}, 2016, pp. 592--600.

\bibitem{kumar2018depth}
H.~Kumar, A.~S. Yadav, S.~Gupta, and K.~Venkatesh, ``Depth map estimation using
  defocus and motion cues,'' \emph{IEEE Trans. Circuits Syst. Video Technol.},
  vol.~29, no.~5, pp. 1365--1379, 2018.

\bibitem{maximov2020focus}
M.~Maximov, K.~Galim, and L.~Leal-Taix{\'e}, ``Focus on defocus: bridging the
  synthetic to real domain gap for depth estimation,'' in \emph{IEEE Conf.
  Comput. Vis. Pattern Recognit. (CVPR)}, 2020, pp. 1071--1080.

\bibitem{lu2021self}
Y.~Lu, G.~Milliron, J.~Slagter, and G.~Lu, ``Self-supervised single-image depth
  estimation from focus and defocus clues,'' \emph{IEEE Robot. Autom. Lett.
  (RAL)}, vol.~6, no.~4, pp. 6281--6288, 2021.

\bibitem{song2020multi}
G.~Song, Y.~Kim, K.~Chun, and K.~M. Lee, ``Multi image depth from defocus
  network with boundary cue for dual aperture camera,'' in \emph{IEEE Int.
  Conf. Acoust. Speech Signal Process. (ICASSP)}, 2020, pp. 2293--2297.

\bibitem{si2023fully}
H.~Si, B.~Zhao, D.~Wang, Y.~Gao, M.~Chen, Z.~Wang, and X.~Li, ``Fully
  self-supervised depth estimation from defocus clue,'' in \emph{IEEE/CVF Conf.
  Comput. Vis. Pattern Recognit. (CVPR)}, 2023, pp. 9140--9149.

\bibitem{wu2024self}
Z.~Wu, Y.~Monno, and M.~Okutomi, ``Self-supervised spatially variant psf
  estimation for aberration-aware depth-from-defocus,'' in \emph{IEEE Int.
  Conf. Acoust. Speech Signal Process. (ICASSP)}, 2024, pp. 2560--2564.

\bibitem{fujimura2024deep}
Y.~Fujimura, M.~Iiyama, T.~Funatomi, and Y.~Mukaigawa, ``Deep depth from focal
  stack with defocus model for camera-setting invariance,'' \emph{Int. J.
  Comput. Vis.}, vol. 132, no.~6, pp. 1970--1985, 2024.

\bibitem{anwar2017depth}
S.~Anwar, Z.~Hayder, and F.~Porikli, ``Depth estimation and blur removal from a
  single out-of-focus image.'' in \emph{Brit. Mach. Vis. Conf. (BMVC)}, vol.~1,
  2017, p.~2.

\bibitem{carvalho2018deep}
M.~Carvalho, B.~Le~Saux, P.~Trouv{\'e}-Peloux, A.~Almansa, and F.~Champagnat,
  ``Deep depth from defocus: how can defocus blur improve 3d estimation using
  dense neural networks?'' in \emph{Eur. Conf. Comput. Vis. (ECCV)}, 2018, pp.
  0--0.

\bibitem{piche2023lens}
D.~Pich{\'e}-Meunier, Y.~Hold-Geoffroy, J.~Zhang, and J.-F. Lalonde, ``Lens
  parameter estimation for realistic depth of field modeling,'' in \emph{IEEE
  Conf. Comput. Vis. Pattern Recognit. (CVPR)}, 2023, pp. 499--508.

\bibitem{wijayasingha2024camera}
L.~Wijayasingha, H.~Alemzadeh, and J.~A. Stankovic, ``Camera-independent single
  image depth estimation from defocus blur,'' in \emph{IEEE Winter Conf. Appl.
  Comput. Vis. (WACV)}, 2024, pp. 3749--3758.

\bibitem{gur2019single}
S.~Gur and L.~Wolf, ``Single image depth estimation trained via depth from
  defocus cues,'' in \emph{IEEE Conf. Comput. Vis. Pattern Recognit. (CVPR)},
  2019, pp. 7683--7692.

\bibitem{li2021single}
Y.~Li, C.~Jung, and J.~Kim, ``Single image depth estimation using edge
  extraction network and dark channel prior,'' \emph{IEEE Access}, vol.~9, pp.
  112\,454--112\,465, 2021.

\bibitem{he2010single}
K.~He, J.~Sun, and X.~Tang, ``Single image haze removal using dark channel
  prior,'' \emph{IEEE Trans. Pattern Anal. Mach. Intell.}, vol.~33, no.~12, pp.
  2341--2353, 2010.

\bibitem{chen2013enhanced}
J.~Chen and L.-P. Chau, ``An enhanced window-variant dark channel prior for
  depth estimation using single foggy image,'' in \emph{IEEE Int. Conf. Image
  Process. (ICIP)}, 2013, pp. 3508--3512.

\bibitem{zhou2023underwater}
J.~Zhou, Q.~Liu, Q.~Jiang, W.~Ren, K.-M. Lam, and W.~Zhang, ``Underwater
  camera: Improving visual perception via adaptive dark pixel prior and color
  correction,'' \emph{Int. J. Comput. Vis.}, pp. 1--19, 2023.

\bibitem{yan2017image}
Y.~Yan, W.~Ren, Y.~Guo, R.~Wang, and X.~Cao, ``Image deblurring via extreme
  channels prior,'' in \emph{IEEE Conf. Comput. Vis. Pattern Recognit. (CVPR)},
  2017, pp. 4003--4011.

\bibitem{pan2017deblurring}
J.~Pan, D.~Sun, H.~Pfister, and M.-H. Yang, ``Deblurring images via dark
  channel prior,'' \emph{IEEE Trans. Pattern Anal. Mach. Intell.}, vol.~40,
  no.~10, pp. 2315--2328, 2017.

\bibitem{cai2020dark}
J.~Cai, W.~Zuo, and L.~Zhang, ``Dark and bright channel prior embedded network
  for dynamic scene deblurring,'' \emph{IEEE Trans. Image Process.}, vol.~29,
  pp. 6885--6897, 2020.

\bibitem{Silberman2012}
N.~Silberman, D.~Hoiem, P.~Kohli, and R.~Fergus, ``Indoor segmentation and
  support inference from rgbd images,'' in \emph{Eur. Conf. Comput. Vis.
  (ECCV)}, 2012, pp. 746--760.

\bibitem{ranftl2020towards}
R.~Ranftl, K.~Lasinger, D.~Hafner, K.~Schindler, and V.~Koltun, ``Towards
  robust monocular depth estimation: Mixing datasets for zero-shot
  cross-dataset transfer,'' \emph{IEEE Trans. Pattern Anal. Mach. Intell.},
  vol.~44, no.~3, pp. 1623--1637, 2020.

\bibitem{haase2020rethinking}
D.~Haase and M.~Amthor, ``Rethinking depthwise separable convolutions: How
  intra-kernel correlations lead to improved mobilenets,'' in \emph{IEEE/CVF
  Conf. Comput. Vis. Pattern Recognit. (CVPR)}, 2020, pp. 14\,600--14\,609.

\bibitem{jin2023depth}
Y.~Jin, M.~Qian, J.~Xiong, N.~Xue, and G.-S. Xia, ``Depth and dof cues make a
  better defocus blur detector,'' in \emph{IEEE Int. Conf. Multimedia Expo
  (ICME)}, 2023, pp. 882--887.

\bibitem{patil2022p3depth}
V.~Patil, C.~Sakaridis, A.~Liniger, and L.~Van~Gool, ``P3depth: Monocular depth
  estimation with a piecewise planarity prior,'' in \emph{IEEE Conf. Comput.
  Vis. Pattern Recognit. (CVPR)}, 2022, pp. 1610--1621.

\bibitem{ke2024repurposing}
B.~Ke, A.~Obukhov, S.~Huang, N.~Metzger, R.~C. Daudt, and K.~Schindler,
  ``Repurposing diffusion-based image generators for monocular depth
  estimation,'' in \emph{IEEE Conf. Comput. Vis. Pattern Recognit. (CVPR)},
  2024, pp. 9492--9502.

\end{thebibliography}

\vfill

\end{document}